%% file: main.tex
\documentclass{article}
\usepackage{lineno}
\usepackage{tencent_ailab_tech_report}
\usepackage[colorlinks = true,
            linkcolor = blue,
            urlcolor  = blue,
            citecolor = blue,
            anchorcolor = blue]{hyperref}   
\usepackage{microtype}
\usepackage{hyperref}
\usepackage{xurl}
\usepackage{enumitem}
\usepackage{multicol}
\usepackage{CJKutf8}
\usepackage{amsmath}
\usepackage{siunitx}
\usepackage{floatflt}
\usepackage{graphicx}
\usepackage{wrapfig}
\usepackage{authblk}
\usepackage{lipsum}
\usepackage{algorithm}
\usepackage{algorithmicx}
\usepackage{algpseudocode}
\usepackage{microtype}
\usepackage{subfigure}
\usepackage{multirow}
\usepackage{pifont}  
\usepackage{amssymb} 
\algnewcommand{\LeftComment}[1]{\Statex \(\triangleright\) #1}
\usepackage{array}
\usepackage{mathtools}
\usepackage{amsthm}

\usepackage[capitalize,noabbrev]{cleveref}
\usepackage{adjustbox} 

\theoremstyle{plain}

\theoremstyle{definition}

\theoremstyle{remark}

\colmfinalcopy

\usepackage[utf8]{inputenc}
\usepackage[T1]{fontenc}
\usepackage{caption} 
\usepackage{adjustbox} 
\usepackage{arydshln}
\usepackage{fontawesome5}

\usepackage{tcolorbox}
\tcbuselibrary{skins,breakable}
\pdfoutput=1

\usepackage{times}
\usepackage{latexsym}
\usepackage{xspace}
\usepackage{natbib}
\usepackage{pgfplots}
\usepackage{tabularx}
\usepackage{subcaption}
\usepackage{bbm}
\usepackage{url}
\usepackage{amsmath} 
\usepackage{makecell}
\usepackage{booktabs}

\DeclareUnicodeCharacter{211D}{\ensuremath{\mathbb{R}}}
\DeclareUnicodeCharacter{2124}{\ensuremath{\mathbb{Z}}}
\DeclareUnicodeCharacter{21A6}{\ensuremath{\mapsto}}

\DeclareUnicodeCharacter{2264}{\ensuremath{\leq}}
\DeclareUnicodeCharacter{2265}{\ensuremath{\geq}}
\DeclareUnicodeCharacter{2192}{\ensuremath{\rightarrow}}
\DeclareUnicodeCharacter{2194}{\ensuremath{\leftrightarrow}}
\DeclareUnicodeCharacter{2227}{\ensuremath{\wedge}}
\DeclareUnicodeCharacter{22A2}{\ensuremath{\vdash}}
\DeclareUnicodeCharacter{2080}{\ensuremath{_0}}
\DeclareUnicodeCharacter{2081}{\ensuremath{_1}}
\DeclareUnicodeCharacter{2082}{\ensuremath{_2}}
\DeclareUnicodeCharacter{2083}{\ensuremath{_3}}
\DeclareUnicodeCharacter{2084}{\ensuremath{_4}}
\DeclareUnicodeCharacter{2085}{\ensuremath{_5}}
\DeclareUnicodeCharacter{2086}{\ensuremath{_6}}
\DeclareUnicodeCharacter{2087}{\ensuremath{_7}}
\DeclareUnicodeCharacter{2088}{\ensuremath{_8}}
\DeclareUnicodeCharacter{2089}{\ensuremath{_9}}
\DeclareUnicodeCharacter{2200}{\ensuremath{\forall}}
\DeclareUnicodeCharacter{2203}{\ensuremath{\exists}}
\DeclareUnicodeCharacter{2228}{\ensuremath{\lor}}
\DeclareUnicodeCharacter{2260}{\ensuremath{\neq}}
\DeclareUnicodeCharacter{230B}{\ensuremath{\rfloor}} 
\DeclareUnicodeCharacter{2223}{\ensuremath{\mid}}    
\DeclareUnicodeCharacter{230A}{\ensuremath{\lfloor}} 
\DeclareUnicodeCharacter{2090}{\ensuremath{_a}}  
\DeclareUnicodeCharacter{2091}{\ensuremath{_e}}  
\DeclareUnicodeCharacter{2092}{\ensuremath{_o}}  
\DeclareUnicodeCharacter{2093}{\ensuremath{_x}}  
\DeclareUnicodeCharacter{2211}{\ensuremath{\sum}}        
\DeclareUnicodeCharacter{2115}{\ensuremath{\mathbb{N}}}  

\DeclareUnicodeCharacter{2208}{\ensuremath{\in}} 

\tcbuselibrary{skins}

\usepackage{color}
\usepackage{xcolor}
\usepackage{soul} 

\definecolor{tred}{RGB}{251, 130, 132}
\definecolor{torange}{RGB}{247, 162, 116}
\definecolor{tyellow}{RGB}{251, 218, 140}
\definecolor{tgreen}{RGB}{127, 204, 181}
\definecolor{tblue}{RGB}{89, 177, 215}
\definecolor{insightblue}{RGB}{162, 210, 255}
\definecolor{questionred}{RGB}{255, 175, 204}
\newcommand{\model}{\textbf{\textsc{EconProver}}\xspace} 
\newcommand{\modelds}{\textbf{\textsc{EconProver-DS}}\xspace} 
\newcommand{\modelgd}{\textbf{\textsc{EconProver-GD}}\xspace} 
\title{\textsc{EconProver}: Towards More Economical Test-Time Scaling for Automated Theorem Proving}

\author{%
Mukai Li\thanks{The work was done when Mukai was interning at Tencent AI Lab.}$^{\phantom{*},\dag,1,2}$,
Linfeng Song\thanks{Correspondence to: Mukai Li \textless limukai.nlp@connect.hku.hk\textgreater, Linfeng Song \textless lfsong@global.tencent.com\textgreater~and~Qi Liu \textless liuqi@hku.hk\textgreater}~~$^{,1}$,
Zhenwen Liang$^{1}$,
Jiahao Xu$^{1}$,
Shansan Gong$^{2}$, \\
\vspace{-5pt}
\textbf{Qi Liu}$^{\dag,2}$,
\textbf{Haitao Mi}$^1$,
\textbf{Dong Yu}$^{1}$
\vspace{10pt}
\\
$^1$Tencent\ \ \ $^2$The University of Hong Kong\\
}

\colmfinalcopy 
\begin{document}
\maketitle
\begin{figure}[h]
\centering
\includegraphics[width=0.80\linewidth]
{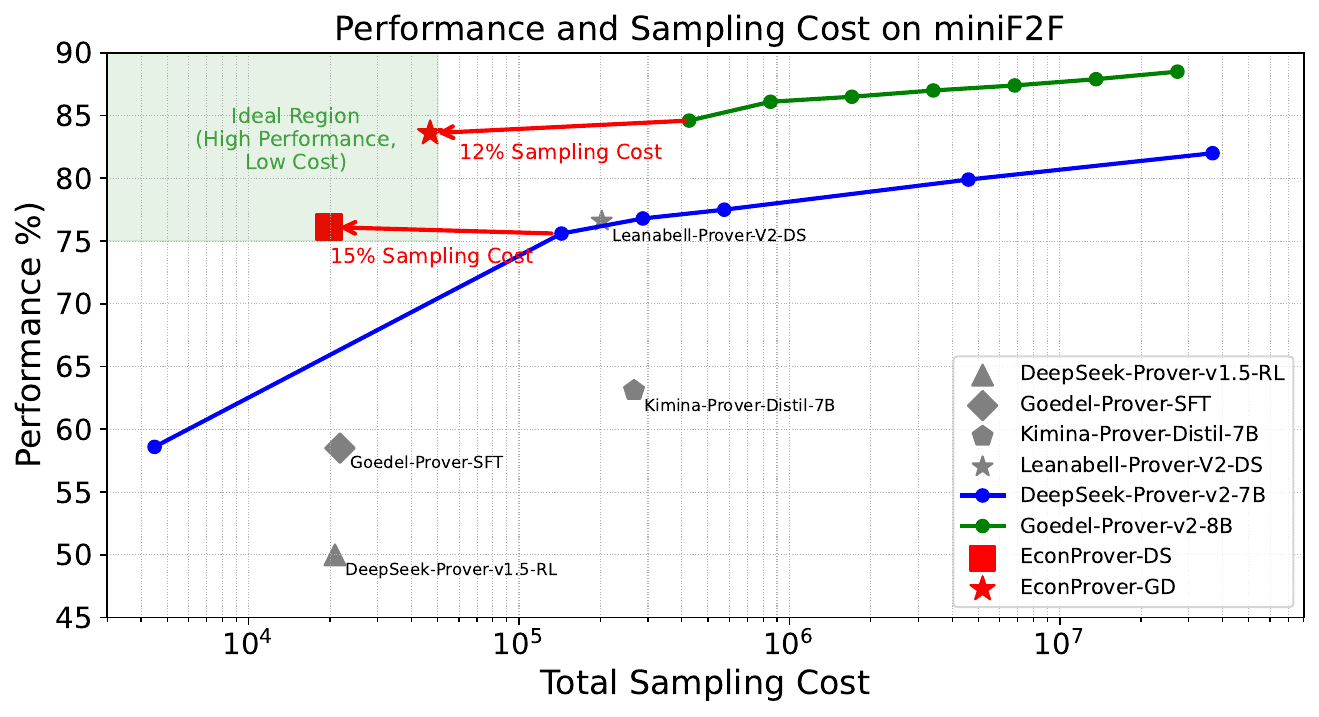}
  \caption{
  Performance-cost curve of open-source ATP models. The horizontal axis shows the total token-level sampling cost (in log scale), and the vertical axis shows the accuracy on miniF2F. 
 Our \modelgd achieves comparable performance while requiring only 12\% of the sampling cost compared to the base model.
  }
  \label{fig:results}
\end{figure}
\begin{abstract}
Large Language Models (LLMs) have recently advanced the field of Automated Theorem Proving (ATP), attaining substantial performance gains through widely adopted test-time scaling strategies, notably reflective Chain-of-Thought (CoT) reasoning and increased sampling passes. However, they both introduce significant computational overhead for inference. Moreover, existing cost analyses typically regulate only the number of sampling passes, while neglecting the substantial disparities in sampling costs introduced by different scaling strategies.
In this paper, we systematically compare the efficiency of different test-time scaling strategies for ATP models and demonstrate the inefficiency of the current state-of-the-art (SOTA) open-source approaches.
We then investigate approaches to significantly reduce token usage and sample passes while maintaining the original performance.
Specifically, we propose two complementary methods that can be integrated into a unified \textbf{EconRL} pipeline for amplified benefits: (1) a dynamic Chain-of-Thought (CoT) switching mechanism designed to mitigate unnecessary token consumption, and (2) Diverse parallel-scaled reinforcement learning (RL) with trainable prefixes to enhance pass rates under constrained sampling passes.
Experiments on miniF2F and ProofNet demonstrate that our \modelgd achieves comparable performance to baseline methods with only 12\% of the computational cost. This work provides actionable insights for deploying lightweight ATP models without sacrificing performance.
\end{abstract}
\input{01intro}

\input{03analysis}
\input{04TTS++}

\input{05experiment}

\input{02related_work}
\input{06conclusion}

\bibliography{custom.bib}
\bibliographystyle{colm2024_conference}

\input{07Appendix}

\end{document}

%% file: 01intro.tex
\section{Introduction}

Automated theorem proving (ATP) aims to automatically generate formal proofs for given mathematical statements.
By transforming natural language statements into theorems in a formal language, e.g., Lean~\citep{moura2021lean} or Isabelle~\citep{wenzel2008isabelle}, and interacting with
the corresponding engine to construct full proofs, an ATP system generates machine-verified
proofs that guarantee strict logical correctness.
The integration of large language models (LLMs) into ATP has marked a paradigm shift in formal mathematics, with systems like DeepSeek-Prover~\citep{ren2025deepseek,lin2025goedelproverv2}, Kimina-Prover~\citep{wang2025kimina}, and Goedel-Prover~\citep{lin2025goedel,lin2025goedelproverv2} achieving unprecedented success rates on challenging benchmarks such as miniF2F \citep{zheng2022miniff} and PutmanBench \citep{tsoukalas2024putnambench}. 
In addition to the improved reasoning capability of the base LLMs, recent advances in ATP predominantly rely on test-time scaling strategies, which can be categorized into two approaches: sequential scaling and parallel scaling.
Inspired from the success of reflective long CoT \citep{deepseekai2025deepseekr1incentivizingreasoningcapability} in general reasoning tasks, \textbf{sequential scaling} augments the proving process with reflective CoT reasoning that mimic how human solves problems.
Typical examples are DeepSeek-Prover-V2 \citep{ren2025deepseek}, which uses natural language sketches to decompose the main theorem into smaller subgoals, and Kimina-Prover \citep{wang2025kimina}, which writes interleaved reasoning and proving blocks to underscore the proving strategy in the macro level.
On the other hand, \textbf{parallel scaling} improves performance by increasing the number of independent sampling passes. Following its widespread success in verifiable tasks like code generation and mathematical reasoning, parallel scaling has been widely adopted by ATP models to boost proving performance.

Despite their apparent effectiveness, 
current state-of-the-art ATP models exhibit significant inefficiencies in their test-time scaling methods.
Sequential scaling techniques typically result in a 10-15 times increase in token usage compared to approaches that do not incorporate reflective CoT reasoning.
Some configurations~\citep{ren2025deepseek} require as much as 8192 proving attempts with over 10,000 tokens per proof attempt, which raises critical questions about deployment feasibility and resource efficiency. The current implementations lack systematic optimization of the fundamental trade-off between computational cost and performance gains. 
Moreover, existing analyses of sampling cost typically focus only on the number of sampling passes, neglecting the impact of token usage and iterative refinement steps. This lack of a unified framework limits our understanding of the true computational trade-offs across different scaling strategies.

To systematically analyze these inefficiencies, we take a unified \textit{sampling cost} metric that sums the total token generation costs across all passes and refinement steps. Our analysis reveals that current methods achieve marginal performance gains at disproportionate computational costs: (1) for problems below IMO difficulty level, models can achieve high performance without complex CoT reasoning, suggesting that applying elaborate chain-of-thought approaches to simpler problems may incur unnecessary computational overhead. (2) Parallel sampling can easily generate redundant proof attempts, causing an early performance plateau.
Motivated by these insights, we propose a unified framework \textbf{EconRL} combining two complementary techniques. First, \textbf{dynamic CoT switching} trains models to autonomously activate extended reasoning only for complex problems via preference learning~\citep{rafailov2024directpreferenceoptimizationlanguage}, maintaining comparable accuracy while reducing token usage to 12\%. Second, \textbf{diverse parallel-scaled RL} employs specialized reasoning heads trained via PPO~\citep{schulman2017proximalpolicyoptimizationalgorithms} on difficulty-partitioned data to improve solution diversity and coverage. Integrating iterative refinement further amplifies the benefits of both techniques.

Our main contributions are as follows:
\begin{itemize}[leftmargin=*]
    \item \textbf{Token-level Cost}: We propose evaluating the inference cost of ATP models in terms of the total number of generated tokens, rather than the number of sampling passes.
    \item \textbf{Inference Efficiency Analysis}: We identify substantial token inefficiency in SOTA provers. For instance, 83\% miniF2F statements below IMO difficulty level can be solved in Non-CoT mode by leading open-source provers (e.g., DeepSeek-Prover-V2), which requires only about 1/10 of the token budget compared to CoT mode. Furthermore, we observe considerable redundancy across different sampled outputs.
    \item \textbf{EconRL}: We proposed a pipelined framework \textbf{EconRL} that stacks two RL tuning stages to improve efficiency. The first stage (\textbf{Dynamic CoT Switching}) conducts preference learning to enable autonomous CoT/Non-CoT mode selection, and the second stage (\textbf{Diverse Parallel-scaled RL}) introduces specialized reasoning heads trained on difficulty-partitioned data, significantly improving solution diversity and parallel scaling efficiency.
    \item \textbf{Practical Efficiency}: Our \modelgd matches the performance of the backbone SOTA open-source ATP model while reducing token consumption to merely 12\%.
\end{itemize}

%% file: 03analysis.tex
\section{Analysis on Test-Time Scaling of ATP models}
\label{sec:q1}
\subsection{Token-Level Sampling Cost Quantification}

We formalize the computational cost of different test-time scaling methods for ATP models through \textit{token-level sampling cost}. The sampling cost is calculated as the sum of initial tokens generated in each pass and the tokens generated during all refinement steps. This framework captures the total inference cost by summing all generation costs across passes and refinement steps, providing a unified metric for evaluating cost-performance trade-offs.

\subsection{Scaling Efficiency Curves\label{sec:scaling_curves}}
\begin{wrapfigure}[26]{r}{0.45\textwidth}
  \vspace{-12pt}
  \centering
\includegraphics[width=\linewidth]{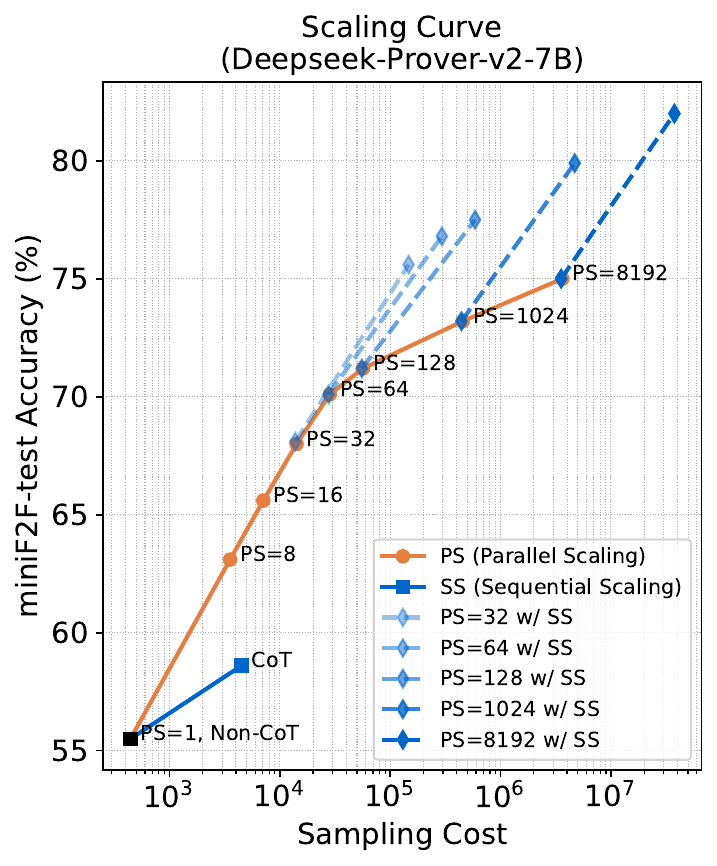}
  \caption{Scaling curve analysis comparing Sequential Scaling (SS), Parallel Scaling (PS), and their combinations (hybrid scaling). The $x$-axis represents the sampling cost in log scale, and the $y$-axis shows the miniF2F-test accuracy. }
  \label{fig:scaling_curve}
\end{wrapfigure}
As shown in Figure~\ref{fig:scaling_curve}, we provide a detailed analysis based on our proposed token-level sampling cost and the model’s performance on the miniF2F ~\citep{zheng2022miniff} benchmark. We examine how model performance varies under parallel scaling (PS) and sequential scaling (SS). Here, SS denotes increasing the token length by enabling reflective CoT mode, while PS=$x$ indicates using $x$ parallel sampling passes. Our analysis reveals three key insights into the efficiency of different scaling strategies:

\paragraph{Sequential scaling demonstrates lower efficiency compared to parallel scaling}
For example, when taking 8 passes (PS=8), parallel scaling already achieves better performance (63.1\% vs 58.6\%) while using fewer tokens (443$\times$8 vs 4,488 tokens) compared to sequential scaling, which enables longer reasoning via reflective CoT.
When further increasing the sampling passes to 16, the performance can reach 65.6\%, another 2.5\% increase, while the sampling cost is not significantly higher than sequential scaling.

\paragraph{Diminishing Returns in Parallel Scaling}
Parallel scaling exhibits clear diminishing returns after increasing passes beyond a certain threshold. While initial increases in passes (from 8 to 32) show substantial gains of 4.9 points, the accuracy improvement becomes increasingly marginal despite exponential growth in computational cost. For instance, doubling passes from 64 to 128 yields only a 1.1\% accuracy gain, indicating that simply scaling up the number of parallel samples can quickly face performance plateau.

\paragraph{Hybrid Scaling Benefits}
Hybrid methods that combine both scaling approaches achieve better efficiency than further increasing parallel passes alone. For example, while increasing PS alone from 32 to 8192 passes yields a 7\% gain, combining PS with sequential scaling (PS=32 w/ SS) achieves comparable improvement (7.5\%) with only around 4\% computational cost than the former (PS=8192). These hybrid methods are simultaneously constrained by the inherent inefficiencies of their components, suggesting that optimizing cost-efficient ATP models requires addressing both the token inefficiency in sequential scaling and the sampling redundancy in parallel scaling simultaneously.

\subsection{Analysis on Scaling methods}

To systematically investigate the inefficiencies identified in our scaling analysis, we design targeted experiments to examine both scaling methods. For sequential scaling, we analyze its effectiveness across problems of varying difficulty levels, revealing unnecessary token consumption for simpler theorems. For parallel scaling, we study the diversity and redundancy of proof attempts under different sampling sizes.
\begin{figure}[!h]
\centering
\includegraphics[width=0.75\columnwidth]{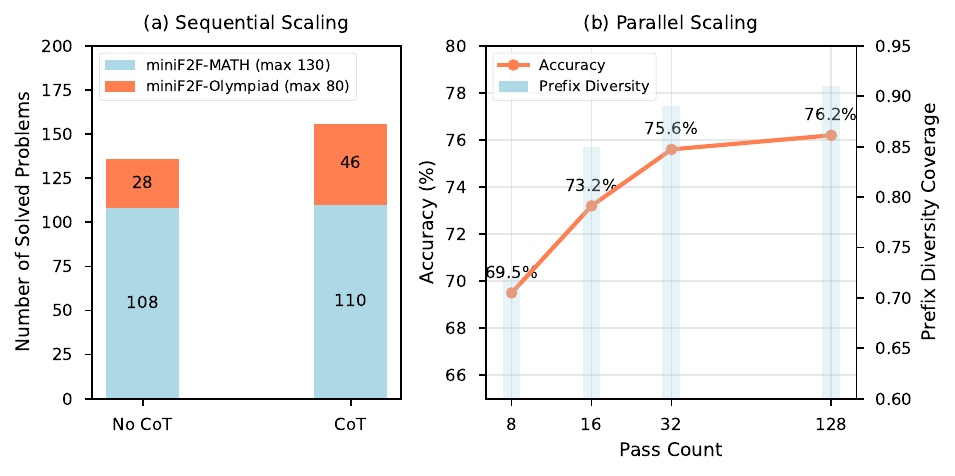}
\caption{Experiments are conducted on DeepSeek-Prover-V2-7B. \textbf{Left}: Sequential scaling improvements are more suitable for solving difficult problems. \textbf{Right}: Parallel scaling accuracy is highly correlated with prefix diversity in generated content.
}
\label{fig:scaling_analysis}
\end{figure}
\paragraph{Sequencial Scaling}
We categorize problems in miniF2F-dev into two difficulty levels: Olympiad-level problems from IMO, AIME, and AMC competitions, and moderate-level problems from MATH algebra and number theory sections. As shown in Figure~\ref{fig:scaling_analysis} left part, while Long CoT shows only modest improvements on moderate-level mathematical problems, its effectiveness becomes particularly pronounced on more challenging Olympiad-level problems. The performance gap between Long CoT and baseline approaches widens significantly for these harder problems. This pronounced contrast suggests that extended reasoning through Long CoT primarily benefits challenging proofs that require sophisticated multi-step logical inference, while simpler problems can often be solved effectively with direct formal proof generation.

\paragraph{Parallel Scaling}
To analyze the diversity of proof attempts, we measure Prefix Diversity Coverage (PDC). For each attempt, we extract the first 20 tokens after the formal statement and break them into 3-grams. We then track how many distinct 3-grams appear as we increase the number of sampled attempts, relative to the full set of 512 attempts. This gives a coverage value between 0 and 1 that grows monotonically with more samples, reflecting how many unique early proof fragments the model explores. Our analysis reveals that this initial proof prefix largely determines the subsequent proof pattern, with the chosen approach persisting throughout generation. As shown in the right part of  Figure~\ref{fig:scaling_analysis}, prefix diversity coverage strongly correlates with proving performance. While significant gains are observed when scaling from 8 to 32 passes (+6.1\% total), further scaling to 128 passes yields minimal returns (+0.6\%) despite 4$\times$ computational cost. This suggests substantial redundancy in exploration beyond 32 passes, motivating strategies that explicitly encourage diverse prefix generation while maintaining lower pass counts.

%% file: 04TTS++.tex
\section{EconRL}
\label{sec:methodology}
Based on our analysis of current scaling inefficiencies, we propose a unified yet lightweight framework \textbf{EconRL} to improve ATP model test-time scaling efficiency from two aspects:
\textbf{dynamic Chain-of-Thought switching} and \textbf{diverse parallel-scaled RL}. The former cuts unnecessary tokens for easy problems, while the latter squeezes more performance out of a small pass budget by encouraging generation diversity. We first introduce each component and then show that their combination yields further gains in the experiments section.
\subsection{Dynamic Chain-of-Thought Switching}
\label{sec:DyCoT}
The inefficiency of uniform CoT application across all problems motivates our development of a dynamic switching mechanism that activates extended reasoning only when necessary. This approach is inspired by recent advances in informal reasoning fields~\citep{lou2025adacotparetooptimaladaptivechainofthought}
but it is specifically adapted for formal theorem proving.

\paragraph{Problem Complexity Classification}
We develop a systematic approach to classify problems based on whether they can be solved by Non-CoT mode. Problems are categorized into two classes.
\begin{itemize}
    \item \textbf{Non-CoT Solvable}: problems correctly solved by base model without CoT reasoning.
    \item \textbf{CoT-dependent}: problems that require extended reasoning for successful proof completion.
\end{itemize}
This classification is performed automatically by running baseline models in the training set and analyzing success patterns. As shown in Figure \ref{fig:scaling_analysis}, roughly 83~\% of the solved problems of Deepseek-Prover-V2-7B below IMO difficulty level are base-solvable, revealing large potential for computational savings.
\paragraph{Preference Learning}
We construct preference pairs $(y_{\text{base}},y_{\text{CoT}})$ for training the switching mechanism with Direct Preference Optimisation (DPO)~\citep{rafailov2024directpreferenceoptimizationlanguage}. The objective is
\begin{equation}
\mathcal{L}_{\text{DPO}} = -\mathbb{E}_{(x,y_w,y_l) \sim D} \Big[ \log \sigma \Big( \beta \log \frac{\pi_\theta(y_w|x)}{\pi_{\text{ref}}(y_w|x)} - \beta \log \frac{\pi_\theta(y_l|x)}{\pi_{\text{ref}}(y_l|x)} \Big) \Big]
\end{equation}
with $\beta=0.1$. The preferred response $y_w$ is the direct proof for base-solvable problems and the full CoT answer for CoT-dependent problems.
The preference construction strategy varies by problem types.
For base-solvable problems: $y_w$ is the direct proof, $y_l$ is the CoT reasoning (to discourage unnecessary reasoning). 
For CoT-dependent problems: $y_w$ stands for CoT reasoning, while $y_l$ is the direct proof (to encourage extended thinking).
This asymmetric preference design enables the model to learn when to apply extended reasoning while avoiding computational waste on simple problems.
\paragraph{Training Data Construction}
We curate a balanced training dataset of 15,000 problem-solution pairs from LeanWorkbook~\citep{ying2025leanworkbooklargescalelean} and Goedel Pset\footnote{https://huggingface.co/datasets/Goedel-LM/Goedel-Pset-v1}. The dataset composition is 40\% base-solvable problems with direct solution preferences and 60\% CoT-dependent problems with extended reasoning preferences.
\paragraph{Prompting Strategy}
Our unified prompt structure, adapted from commonly used prompts in current ATP models to ensure seamless distribution alignment between CoT and Non-CoT modes, enables the model to dynamically determine the appropriate reasoning mode without explicit complexity indicators.
\begin{figure*}[h]
\begin{tcolorbox}[colback=gray!10, title=Dynamic CoT Prompt Template]
\small
\texttt{Complete the following Lean 4 code, thinking step by step if the problem requires careful reasoning:}\\
\texttt{<problem\_statement>}\\
\texttt{/- The model will autonomously choose:}\\
\texttt{   Option 1: Direct formal proof generation} \\
\texttt{   Option 2: Informal reasoning followed by formal proof -/}
\end{tcolorbox}
\end{figure*}
During inference, the model learns to assess problem complexity implicitly through attention patterns learned during DPO training. The model develops internal representations that correlate with problem difficulty, enabling automatic mode selection without external complexity signals.
\subsection{Diverse Parallel-scaled RL}
\label{Parallel-RL}
As discussed in our analysis section, parallel scaling can consistently improve performance by increasing the number of independent proof attempts. However, the marginal gains diminish rapidly at higher pass counts, as many attempts become redundant and repeatedly explore similar solution paths, leading to significant computational overhead for small improvements. To address this inefficiency, we aim to make each attempt more diverse, thereby increasing the efficiency of parallel exploration under a limited sampling budget. Concretely, we learn $n$ specialized heads, implemented as lightweight prefix embeddings, that cooperate to cover the proof space more effectively.
\paragraph{Difficulty-aware Partitioning}
Our key insight is that difficulty-aware data partitioning enables learning specialized prefix tokens that generate more diverse proof attempts. Specifically, we measure problem difficulty by running the base prover for 32 attempts and recording the success count $c(x)$ for each problem. After sorting problems by $c(x)$, we partition the dataset into $n$ difficulty-based bins $B_1,\ldots,B_n$ from easy to hard. For each bin $B_i$, we construct a training shard $S_i$ containing 50\% problems from $B_i$ and 50\% sampled uniformly from other bins, ensuring both specialization and generalization. Each shard is paired with a dedicated prefix (head) $P_i$ that learns difficulty-specific proof strategies. Through extensive experiments shown in \S\ref{Ablation_on_para}, we found that this difficulty-aware partitioning significantly outperforms naive diversification approaches like random heads, which yield only marginal improvements. Based on empirical validation, we set $n=8$ heads as the optimal configuration for balancing diversity and computational cost.

\paragraph{Learnable Parallel Scaling via Trainable 
Prefixes}
We employ Proximal Policy Optimization (PPO)~\citep{schulman2017proximalpolicyoptimizationalgorithms} to train each prefix (head) \emph{independently}, without joint optimization across heads. For each prefix, the training objective is defined as follows. For a given proof attempt, the model receives a reward of 1 if the generated proof is correct, and 0 otherwise.  
Notably, each prefix is optimized separately, and there is no joint or coordinated update across the eight heads. This independent optimization framework allows each prefix to specialize in different proof strategies, thereby increasing the diversity and coverage of the parallel search process.
\paragraph{Uniform Allocation at Inference}
During evaluation we distribute the sampling budget evenly across the heads (e.g., Pass@16~$\Rightarrow$~$16/n$ calls per head). This zero-cost scheduler retains the learned diversity and avoids expensive per-problem head selection.


%% file: 05experiment.tex
\section{Experiments}
\label{sec:exp}
We conduct extensive experiments to evaluate the efficiency improvements of our proposed \textbf{EconRL} for ATP models. Our experiments demonstrate that dynamic CoT switching and diverse parallel-scaled RL not only reduce computational costs individually, but also exhibit strong synergistic effects when combined with various backbone models. Furthermore, we show that these efficiency optimizations remain effective when integrated with advanced techniques like iterative refinement, enabling substantial cost reductions even in high-performance settings.
\subsection{Experimental Setup}

\paragraph{Models}
Our \modelds and \modelgd are based on the following two representative open-source ATP models. 
\textbf{DeepSeek-Prover-V2-7B}~\citep{ren2025deepseek} is built upon DeepSeek-Prover-V1.5-Base and features an extended context length of up to 32K tokens, leveraging Lean 4 integration for formal verification and synthetic data generation via DeepSeek-V3 decomposition. 
\textbf{Goedel-Prover-V2-8B}~\citep{lin2025goedelproverv2} utilizes expert iteration training with dual formalizers to translate informal statements into Lean 4, progressively expanding its proof dataset by solving previously unprovable theorems. We adopt \textbf{EconRL} (\S\ref{sec:methodology}), where the base model is initially trained with the dynamic CoT switching method (\S\ref{sec:DyCoT}) and subsequently with diverse parallel-scaled RL training (\S\ref{Parallel-RL}). For CoT and Non-CoT mode, we adopt prompt used in the original paper. For iterative refinement, we follow the setting in Goedel-Prover-V2.
\paragraph{Baselines}
\textbf{Goedel-Prover-SFT}~\citep{lin2025goedelproverv2} is supervised fine-tuning upon DeepSeek-Prover-V1.5-Base on Goedel-Pset-v1-solved.
\textbf{Kimina-Prover-Preview-Distill-7B}~\citep{wang2025kimina} combines symbolic reasoning with neural guidance through interactive proof-state pruning. \textbf{Leanabell-Prover-V2-DS}~\citep{ji2025leanabellproverv2verifierintegratedreasoningformal} is developed by further finetuning DeepSeek-Prover-V2 using RL with feedback from Lean 4.
\paragraph{Datasets}
We evaluate our approach on two standard theorem proving benchmarks: \textbf{miniF2F-test}~\citep{zheng2022miniff} and \textbf{ProofNet-test}~\citep{azerbayev2023proofnetautoformalizingformallyproving}. We follow recent work and use the revised version of miniF2F provided by KiminaProver~\citep{wang2025kimina}, which corrects several errors in the original dataset. MiniF2F consists of 488 problems in Lean, drawn from high-school level competitions such as the AMC, AIME, and the IMO. The benchmark covers core areas of elementary mathematics (e.g., algebra, number theory, and induction) and is split evenly into miniF2F-valid and miniF2F-test, each containing 244 problems. We reserve the test split for evaluation.
For ProofNet-test, we follow the data sources and splits used in DeepSeek-Prover-V2~\citep{ren2025deepseek}. In particular, the dataset consists of 371 problems in Lean 3, drawn from a range of popular undergraduate pure mathematics textbooks, covering topics such as real and complex analysis, linear algebra, abstract algebra, and topology. Following DeepSeek-Prover-V2, we use the Lean 4 translation version of the ProofNet test set, which consists of 186 problems.
\subsection{Main Performance and Efficiency}
\begin{table*}[!h]
\centering
\caption{Accuracy and efficiency comparison on miniF2F-test and ProofNet-test. Sampling cost is calculated by summing all generation costs across passes and refinement steps. Rel. denotes the relative sampling token cost, expressed as an approximate multiple of the same model’s non-CoT baseline.}
\label{tab:results}
\resizebox{\linewidth}{!}{
\begin{tabular}{lrcrrcr}
\toprule
\textbf{Model / Setting} & \multicolumn{3}{c}{\textbf{miniF2F}} & \multicolumn{3}{c}{\textbf{ProofNet}} \\
\cmidrule(lr){2-4} \cmidrule(lr){5-7}
& \textbf{Cost} & \textbf{Acc (\%)} & \textbf{Rel.} & \textbf{Cost} & \textbf{Acc (\%)} & \textbf{Rel.} \\
\midrule
\textit{Goedel-Prover-SFT} & 0.7k$\times$32 & 58.5 & -- & 0.7k$\times$32 & 15.6 & -- \\
\textit{Kimina-Prover-distil-7B} & 8.3k$\times$32 & 63.1 & -- & -- & -- & -- \\
\textit{Leanabell-Prover-V2-DS} \
&6.3k$\times$32  &76.6 & -- & 8.9k$\times$32 & 23.7 & --\\
\midrule
\textit{Deepseek-Prover-V2-7B} \\
\quad Non-CoT mode & 0.4k$\times$32 & 68.1 & 1$\times$ & 0.5k$\times$32 & 21.5 & 1$\times$ \\
\quad CoT mode & 4.5k$\times$32 & 75.8 & 10$\times$ & 7.0k$\times$32 & 23.1 & 10$\times$ \\
\quad \modelds & 1.2k$\times$16 & 76.2 & 1.5$\times$ & 2.2k$\times$16 & 23.1 & 2$\times$ \\
\midrule
\textit{Goedal-Prover-V2-8B} \\
\quad Non-CoT mode & 0.5k$\times$32 & 75.8 & 1$\times$ & 0.5k$\times$32 & 24.7 & 1$\times$ \\
\quad CoT mode & 12.3k$\times$32 & 84.4 & 25$\times$ & 15.2k$\times$32  & 28.5 & 30$\times$\\
\quad \modelgd & 2.9k$\times$16 & 84.0 & 3$\times$ & 3.1k$\times$16 &28.0 & 3$\times$ \\
\hdashline
\quad CoT mode + IR & 12.3k+3.7k$\times$2)$\times$32 & 86.0 & 40$\times$ & -- & -- & -- \\
\quad \modelgd + IR & (2.9k+3.6k$\times$2)$\times$16 & 86.0 & 10$\times$ & -- & -- & -- \\
\bottomrule
\end{tabular}}
\end{table*}
\paragraph{Efficient Performance through Hybrid Approach}

As shown in Table \ref{tab:results}, our experimental results demonstrate the remarkable efficiency of our hybrid approach in automated theorem proving. By combining dynamic CoT switching with diverse parallel-scaled RL, our \modelds achieves performance comparable to full CoT methods while significantly reducing computational overhead. Leanabell-Prover-V2-DS, which is also using DeepSeek-Prover-V2 as the base model, achieves slightly higher performance, but the sampling cost increased nearly 50\% than its base model. 
In contrast, our method maintains the high accuracy of CoT-based approaches (achieving 76.2\% on miniF2F-test) while requiring only 15\% of the token usage compared to standard CoT implementations. This efficiency gain is achieved through intelligent resource allocation, where the system dynamically determines when to employ detailed reasoning steps and optimizes exploration through specialized reasoning heads.

\paragraph{Strong Generalization Across ATP Models}
The effectiveness of our \textbf{EconRL} shows strong generalization across different model architectures and benchmarks. When applied to both DeepSeek-Prover-V2 and Goedel-Prover-V2, our \modelds and \modelgd consistently delivers substantial efficiency improvements while maintaining competitive performance. On miniF2F-test, the approach achieves 76.2\% accuracy with DeepSeek-Prover-V2 and 84.0\% with Goedel-Prover-V2, demonstrating its adaptability across different foundation models. Similarly, on ProofNet, our method maintains the strong performance of baseline models while significantly reducing computational costs, highlighting the broad applicability of our efficiency optimizations across diverse theorem-proving tasks.

\paragraph{Compatibility with Iterative Refinement}
Recent advances in ATP systems have demonstrated the effectiveness of iterative refinement (IR) in improving proving performance. As shown in Table~\ref{tab:results}, Goedel-Prover-V2 with IR achieves 86.0\% accuracy, but at the cost of substantially increased token usage (60 times compared to baseline). When combined with IR, our \textbf{EconRL} optimizations maintain their effectiveness, reducing the token overhead by 75\% while preserving the 86.0\% accuracy. This demonstrates that our method can effectively optimize computational costs even in advanced proving frameworks that push the boundaries of performance.
\subsection{Ablation on Dynamic CoT Switching}
\begin{wraptable}{r}{0.55\textwidth}
\vspace{-1.2em}
\centering
\caption{Dynamic CoT Switching on DeepSeek-Prover-V2}
\label{tab:dpo_results}
\resizebox{\linewidth}{!}{
\begin{tabular}{lcrr}
\toprule
\textbf{Configuration} & \textbf{Pass@32} & \textbf{\#Avg} & \textbf{CoT Rate}  \\
 & \textbf{(\%)} & & \textbf{(\%)}~~~ \\
\midrule
Non-CoT (Baseline) & 68.0 & 443 & 0.0  \\
Direct Prompt & 71.7 & 1,433 & 18.1   \\
\textbf{Dynamic CoT Switching} & 75.4 & \textbf{1,186} & \textbf{14.8}   \\
Full CoT (Always) & \textbf{75.8} & 4,488 & 100.0 \\
\bottomrule
\end{tabular}}
\vspace{1em}
\end{wraptable}
To better understand the effectiveness of our Dynamic CoT Switching approach, we conduct detailed ablation studies using DeepSeek-Prover-V2 as the base model. As shown in Table~\ref{tab:dpo_results}, our method achieves remarkable efficiency while maintaining high performance. Compared to Full CoT, Dynamic CoT Switching achieves 99.7\% of the accuracy (75.4\% vs 75.8\%) while requiring only 15\% of the token usage (1,186 vs 4,488.3 average tokens). When compared to the Direct Prompt baseline, our method shows significantly better performance (75.4\% vs 71.7\%) with better token efficiency. Moreover, relative to Non-CoT baseline, Dynamic CoT Switching substantially improves accuracy (75.4\% vs 68.0\%) while keeping the computational overhead at a reasonable level through intelligent CoT rate control (14.8\%).
\subsection{Ablation on Diverse Parallel-scaled RL}
\label{Ablation_on_para}
\begin{wraptable}{r}{0.55\textwidth}
\vspace{-1.2em}
\centering
\caption{Random vs. our difficulty-aware grouping based on DeepSeek-Prover-V2 Non-CoT mode. ``Cover@$k$'' represents ``Prefix Diversity Coverage@$k$''.}
\label{tab:difficulty_grouping}
\begin{tabular}{lcccc}
\toprule
\textbf{Method} & \textbf{\#Heads} & \textbf{Pass@8} & \textbf{Pass@16} &\textbf{Diversity}\\
 &  &\textbf{(\%)} & \textbf{(\%)} & \textbf{Cover@8} \\
 \midrule
Baseline & - & 63.1 & 65.6  & 74.3\\
\midrule
& 4 &65.6 &66.8  &74.5\\
Random & 8 & 66.8 & 68.9 &74.8\\
& 16 &66.8 &68.9 & 75.1\\
\midrule
& 4 & 67.2 & 69.3 & 82.3\\
Ours & 8 & \textbf{68.9} & \textbf{70.5} &84.1\\
& 16 & \textbf{68.9} & 70.1 &\textbf{84.5}\\
\bottomrule
\end{tabular}
\vspace{-2em}
\end{wraptable}
\paragraph{Number of Parallel Heads} 
As shown in Table~\ref{tab:difficulty_grouping}, we first investigate the impact of different numbers of parallel heads ($n \in \{4,8,16\}$). With a random grouping strategy, increasing heads from 4 to 8 improves Pass@16 from 66.8\% to 68.9\%, while further increasing to 16 heads yields no additional gains (68.9\%). 
\paragraph{Difficulty-aware Data Grouping}
We then examine the effectiveness of our difficulty-aware grouping strategy compared to random grouping. With the same number of heads ($n=8$), difficulty-aware grouping significantly outperforms random grouping on both CoT and Non-CoT settings. This improvement is consistent across different head numbers, demonstrating that intelligently grouping training data based on problem difficulty helps specialized heads better capture different reasoning patterns. 
\paragraph{Results}
\begin{wraptable}{r}{0.55\textwidth}
\vspace{-0em}
\centering
\caption{Random vs diverse parallel-scaled RL on CoT mode with larger sampling passes.}
\label{tab:parallel_results}
\begin{tabular}{lcc}
\toprule
\textbf{Configuration} & \textbf{Pass@16} & \textbf{Pass@32}\\
   &(\%) & (\%) \\
\midrule
\multicolumn{3}{l}{\textit{DeepSeek-Prover-V2 }}\\
CoT mode & 73.0 & 75.8\\
~~~+ 8 heads (random) & 73.4 & 75.8\\
~~~+ 8 heads (diff-aware) & \textbf{75.0} & \textbf{76.2}\\
\bottomrule
\end{tabular}
\vspace{0em}
\end{wraptable}
Table~\ref{tab:parallel_results} demonstrates the effectiveness of our diverse parallel-scaled RL approach across different settings. In the Non-CoT setting, our method significantly improves performance with fewer passes, achieving 70.5\% at Pass@16 compared to the baseline's 65.6\%. Similarly, with CoT enabled, we observe consistent gains, reaching 75.0\% at Pass@16 versus 73.0\% baseline. Notably, our approach shows stronger benefits at lower pass counts (Pass@16) compared to higher ones (Pass@32), indicating its effectiveness in maximizing performance when computational resources are constrained. The results also show that simply using random heads provides minimal benefits, highlighting the importance of our difficulty-aware training strategy.

%% file: 02related_work.tex
\section{Related Work}
\subsection{Formal Theorem Proving via whole proof generation} 
Inspired by the success of end-to-end reasoning in informal LLMs, recent work in formal theorem proving has explored generating entire proofs in one forward pass. In environments like Lean \citep{moura2021lean}, a model's output can be formally verified, avoiding the logically flawed yet fluent arguments often seen in free-form reasoning. Several large formal models such as \textbf{DeepSeek-Prover}~\citep{xin2024deepseekproveradvancingtheoremproving}, \textbf{Goedel-Prover}~\citep{lin2025goedel}, demonstrate that whole-proof generation can produce correct Lean proofs that pass the Lean checker. This paradigm retains the global coherence and simplicity of sequence generation while ensuring rigor through verification. Recently, with the rise of long chain-of-thought (CoT) reasoning \citep{openai2024openaio1card,deepseekai2025deepseekr1incentivizingreasoningcapability}, ATP models have also started adopting long CoT generation before producing Lean code \citep{xin2024deepseekproverv15harnessingproofassistant,wang2025kimina}. While these methods can further improve performance on theorem proving tasks, one challenge in this approach is the length of generated proofs: using explicit chain-of-thought (CoT) reasoning can make proof outputs significantly longer \citep{ren2025deepseek}, inflating token usage. Our work follows this whole-proof direction, further \textbf{reducing the token length} by allowing the model to invoke detailed reasoning only when necessary. This yields more compact proofs and directly cuts down token consumption.
\subsection{Formal Theorem Proving with Verifier Feedback} 
Research leveraging verifier feedback in formal theorem proving has evolved along two main directions: stepwise search-based provers that construct proofs incrementally with per-step verification, and whole-proof refinement approaches that generate complete proofs and iteratively improve them based on verifier feedback.

In the first direction, stepwise search-based provers explore a tree of proof states, expanding one step at a time and using the proof assistant to check each step. Systems such as \textbf{InternLM2.5-StepProver} \citep{wu2024internlm}, \textbf{HunyuanProver} \citep{li2024hunyuan,liang2025mps}, \textbf{DeepSeek-Prover-V1.5} \citep{xin2024deepseekproverv15harnessingproofassistant}, and \textbf{BFS-Prover} \citep{xin2025bfs} employ best-first or Monte Carlo tree search to try many possible tactic sequences. \textbf{DSP+} \citep{cao2025dsp} and \textbf{DRP} \citep{liang2025towards} use a powerful general LLM to decompose~\citep{Zhao2024SubgoalXLSE} the problem and then delegate fine-grained steps to a formal prover. This approach improves success rates by iteratively repairing failures, which means if a step is invalid, the model backtracks and tries a different branch – thereby eventually discovering a valid proof if one exists. However, the exhaustive search comes at the cost of \textbf{huge computational overhead}: the model must evaluate a large number of partial proofs, making these methods extremely resource-intensive.

The second direction focuses on iterative refinement, where models first generate complete proofs and then iteratively refine them based on verifier feedback. Notable examples include \textbf{Seed-Prover}~\citep{chen2025seedproverdeepbroadreasoning}, \textbf{Goedel-Prover-V2}~\citep{lin2025goedelproverv2} and \textbf{StepFun-Prover}~\citep{shang2025stepfunprover}, which use sophisticated refinement strategies to improve initially generated proofs. At the extreme end, \textbf{Seed-Prover} \citep{chen2025seedproverdeepbroadreasoning} combines multi-stage reinforcement learning, an agentic search strategy, and extensive test-time exploration to reach \emph{IMO medal-level} performance, fully solving 5 of 6 problems at IMO~2025 and essentially saturating the miniF2F benchmark. This represents a remarkable leap in capability, but it demands enormous computational resources (dozens of refinement iterations). 
While iterative refinement approaches achieve strong performance, they require substantial computational resources for multiple refinement iterations. Our approach can be effectively combined with these refinement methods while significantly reducing the computational overhead. 
\subsection{Dynamic Chain-of-Thought}
Recent work has explored dynamic Chain-of-Thought (CoT) prompting methods that let an LLM decide on-the-fly whether to engage in multi-step reasoning for a given query, in order to balance accuracy and token cost. For example, \textbf{AdaCoT} \citep{Lou2025AdaCoT} uses a reinforcement learning policy to trigger CoT only on complex questions, formalizing a Pareto-optimal trade-off between reasoning performance and inference overhead. Similarly, \textbf{L1}~\citep{Aggarwal2025Controlling} trains an RL-based controller that enforces chain-of-thought length constraints, allowing the model to trade off minimal accuracy loss for significant token savings. Other approaches adopt utility or difficulty-based triggers: \textbf{DynaThink} \citep{Pan2024DynaThink} dynamically chooses between a ``fast'' direct answer mode and a ``slow'' multi-step mode based on the model’s confidence in the query, while \textbf{DAST} \citep{Shen2025DAST} adjusts the CoT reasoning depth according to predicted problem difficulty, cutting unnecessary steps without harming performance on challenging queries. All these adaptive CoT frameworks seek to maximize reasoning accuracy on difficult inputs while avoiding verbose reasoning on trivial ones. To our knowledge, however, this paper is the first to apply dynamic CoT triggering in the context of formal theorem proving, where tightly controlling the proof generation length is crucial.

%% file: 06conclusion.tex
\section{Conclusion and Future Work}
In this work, we presented a comprehensive approach \textbf{EconRL} to improve the test-time scaling efficiency of ATP models through two complementary techniques: dynamic Chain-of-Thought switching and diverse parallel-scaled reinforcement learning. Our dynamic CoT mechanism significantly reduces token overhead by selectively applying extended reasoning, while our parallel RL strategy improves performance with limited sampling budgets through specialized proof heads. Together, our \model demonstrate that substantial efficiency gains are possible without sacrificing proving capability.
Looking ahead, we plan to extend our efficiency-focused methodology to iterative refinement frameworks. By integrating dynamic CoT switching and parallel RL training into the iterative proving process, we aim to optimize ATP model inference costs along all the three dimensions. This holistic approach promises to reduce computational overhead across the entire proving pipeline while maintaining or improving proving performance. Such comprehensive optimization will be crucial for making advanced ATP systems more practical and accessible for real-world applications.

%% file: 07appendix.tex
\newpage
\appendix

%% file: main.bbl
\begin{thebibliography}{31}
\providecommand{\natexlab}[1]{#1}
\providecommand{\url}[1]{\texttt{#1}}
\expandafter\ifx\csname urlstyle\endcsname\relax
  \providecommand{\doi}[1]{doi: #1}\else
  \providecommand{\doi}{doi: \begingroup \urlstyle{rm}\Url}\fi

\bibitem[Aggarwal \& Welleck(2025)Aggarwal and Welleck]{Aggarwal2025Controlling}
Pranjal Aggarwal and Sean Welleck.
\newblock Controlling how long a reasoning model thinks with reinforcement learning.
\newblock \emph{arXiv preprint arXiv:2503.04697}, 2025.

\bibitem[Azerbayev et~al.(2023)Azerbayev, Piotrowski, Schoelkopf, Ayers, Radev, and Avigad]{azerbayev2023proofnetautoformalizingformallyproving}
Zhangir Azerbayev, Bartosz Piotrowski, Hailey Schoelkopf, Edward~W. Ayers, Dragomir Radev, and Jeremy Avigad.
\newblock Proofnet: Autoformalizing and formally proving undergraduate-level mathematics, 2023.
\newblock URL \url{https://arxiv.org/abs/2302.12433}.

\bibitem[Cao et~al.(2025)Cao, Song, Li, Le, Zhang, Xue, and Yang]{cao2025dsp}
Chenrui Cao, Liangcheng Song, Zenan Li, Xinyi Le, Xian Zhang, Hui Xue, and Fan Yang.
\newblock Reviving {DSP} for advanced theorem proving in the era of reasoning models (dsp+).
\newblock \emph{CoRR}, abs/2506.11487, 2025.
\newblock URL \url{https://arxiv.org/abs/2506.11487}.

\bibitem[Chen et~al.(2025)Chen, Gu, Huang, Huang, Jiang, Jie, Jin, Jin, Li, Ma, Ren, Shen, Shi, Sun, Sun, Wang, Wang, Wang, Wei, Wei, Wu, Wu, Xia, Xin, Yang, Ying, Yuan, Yuan, Zhan, Zhang, Zhang, Zhang, Zhao, Zhao, Zhou, and Zhu]{chen2025seedproverdeepbroadreasoning}
Luoxin Chen, Jinming Gu, Liankai Huang, Wenhao Huang, Zhicheng Jiang, Allan Jie, Xiaoran Jin, Xing Jin, Chenggang Li, Kaijing Ma, Cheng Ren, Jiawei Shen, Wenlei Shi, Tong Sun, He~Sun, Jiahui Wang, Siran Wang, Zhihong Wang, Chenrui Wei, Shufa Wei, Yonghui Wu, Yuchen Wu, Yihang Xia, Huajian Xin, Fan Yang, Huaiyuan Ying, Hongyi Yuan, Zheng Yuan, Tianyang Zhan, Chi Zhang, Yue Zhang, Ge~Zhang, Tianyun Zhao, Jianqiu Zhao, Yichi Zhou, and Thomas~Hanwen Zhu.
\newblock Seed-prover: Deep and broad reasoning for automated theorem proving, 2025.
\newblock URL \url{https://arxiv.org/abs/2507.23726}.

\bibitem[DeepSeek-AI(2025)]{deepseekai2025deepseekr1incentivizingreasoningcapability}
DeepSeek-AI.
\newblock Deepseek-r1: Incentivizing reasoning capability in llms via reinforcement learning, 2025.
\newblock URL \url{https://arxiv.org/abs/2501.12948}.

\bibitem[Ji et~al.(2025)Ji, Liu, Wang, Zhang, Yue, Shi, Sun, Zhang, Zhou, and Gai]{ji2025leanabellproverv2verifierintegratedreasoningformal}
Xingguang Ji, Yahui Liu, Qi~Wang, Jingyuan Zhang, Yang Yue, Rui Shi, Chenxi Sun, Fuzheng Zhang, Guorui Zhou, and Kun Gai.
\newblock Leanabell-prover-v2: Verifier-integrated reasoning for formal theorem proving via reinforcement learning, 2025.
\newblock URL \url{https://arxiv.org/abs/2507.08649}.

\bibitem[Li et~al.(2024)Li, Du, Song, Li, Wang, Yang, and Mi]{li2024hunyuan}
Yang Li, Dong Du, Linfeng Song, Chen Li, Weikang Wang, Tao Yang, and Haitao Mi.
\newblock Hunyuanprover: A scalable data synthesis framework and guided tree search for automated theorem proving.
\newblock \emph{CoRR}, abs/2412.20735, 2024.
\newblock URL \url{https://arxiv.org/abs/2412.20735}.

\bibitem[Liang et~al.(2025{\natexlab{a}})Liang, Song, Li, Yang, Zhang, Mi, and Yu]{liang2025mps}
Zhenwen Liang, Linfeng Song, Yang Li, Tao Yang, Feng Zhang, Haitao Mi, and Dong Yu.
\newblock Mps-prover: Advancing stepwise theorem proving by multi-perspective search and data curation.
\newblock \emph{arXiv preprint arXiv:2505.10962}, 2025{\natexlab{a}}.

\bibitem[Liang et~al.(2025{\natexlab{b}})Liang, Song, Li, Yang, Zhang, Mi, and Yu]{liang2025towards}
Zhenwen Liang, Linfeng Song, Yang Li, Tao Yang, Feng Zhang, Haitao Mi, and Dong Yu.
\newblock Towards solving more challenging imo problems via decoupled reasoning and proving.
\newblock \emph{arXiv preprint arXiv:2507.06804}, 2025{\natexlab{b}}.

\bibitem[Lin et~al.(2025{\natexlab{a}})Lin, Tang, Lyu, Wu, Lin, Yang, Li, Xia, Chen, Arora, et~al.]{lin2025goedel}
Yong Lin, Shange Tang, Bohan Lyu, Jiayun Wu, Hongzhou Lin, Kaiyu Yang, Jia Li, Mengzhou Xia, Danqi Chen, Sanjeev Arora, et~al.
\newblock Goedel-prover: A frontier model for open-source automated theorem proving.
\newblock \emph{arXiv preprint arXiv:2502.07640}, 2025{\natexlab{a}}.

\bibitem[Lin et~al.(2025{\natexlab{b}})Lin, Tang, Lyu, Yang, Chung, Zhao, Jiang, Geng, Ge, Sun, Wu, Gesi, Acuna, Yang, Lin, Choi, Chen, Arora, and Jin]{lin2025goedelproverv2}
Yong Lin, Shange Tang, Bohan Lyu, Ziran Yang, Jui-Hui Chung, Haoyu Zhao, Lai Jiang, Yihan Geng, Jiawei Ge, Jingruo Sun, Jiayun Wu, Jiri Gesi, David Acuna, Kaiyu Yang, Hongzhou Lin, Yejin Choi, Danqi Chen, Sanjeev Arora, and Chi Jin.
\newblock Goedel-prover-v2: The strongest open-source theorem prover to date, 2025{\natexlab{b}}.

\bibitem[Lou et~al.(2025{\natexlab{a}})Lou, Sun, Liang, Qu, Shen, Wang, Li, Yang, and Wu]{Lou2025AdaCoT}
Chenwei Lou, Zewei Sun, Xinnian Liang, Meng Qu, Wei Shen, Wenqi Wang, Yuntao Li, Qingping Yang, and Shuangzhi Wu.
\newblock Adacot: Pareto-optimal adaptive chain-of-thought triggering via reinforcement learning.
\newblock \emph{arXiv preprint arXiv:2505.11896}, 2025{\natexlab{a}}.

\bibitem[Lou et~al.(2025{\natexlab{b}})Lou, Sun, Liang, Qu, Shen, Wang, Li, Yang, and Wu]{lou2025adacotparetooptimaladaptivechainofthought}
Chenwei Lou, Zewei Sun, Xinnian Liang, Meng Qu, Wei Shen, Wenqi Wang, Yuntao Li, Qingping Yang, and Shuangzhi Wu.
\newblock Adacot: Pareto-optimal adaptive chain-of-thought triggering via reinforcement learning, 2025{\natexlab{b}}.
\newblock URL \url{https://arxiv.org/abs/2505.11896}.

\bibitem[Moura \& Ullrich(2021)Moura and Ullrich]{moura2021lean}
Leonardo~de Moura and Sebastian Ullrich.
\newblock The lean 4 theorem prover and programming language.
\newblock In \emph{International Conference on Automated Deduction}, pp.\  625--635. Springer, 2021.

\bibitem[OpenAI(2024)]{openai2024openaio1card}
OpenAI.
\newblock Openai o1 system card, 2024.
\newblock URL \url{https://arxiv.org/abs/2412.16720}.

\bibitem[Pan et~al.(2024)Pan, Zhang, Zhang, Liu, Wang, and Li]{Pan2024DynaThink}
Jiabao Pan, Yan Zhang, Chen Zhang, Zuozhu Liu, Hongwei Wang, and Haizhou Li.
\newblock Dynathink: Fast or slow? a dynamic decision-making framework for large language models.
\newblock \emph{arXiv preprint arXiv:2407.01009}, 2024.

\bibitem[Rafailov et~al.(2024)Rafailov, Sharma, Mitchell, Ermon, Manning, and Finn]{rafailov2024directpreferenceoptimizationlanguage}
Rafael Rafailov, Archit Sharma, Eric Mitchell, Stefano Ermon, Christopher~D. Manning, and Chelsea Finn.
\newblock Direct preference optimization: Your language model is secretly a reward model, 2024.
\newblock URL \url{https://arxiv.org/abs/2305.18290}.

\bibitem[Ren et~al.(2025)Ren, Shao, Song, Xin, Wang, Zhao, Zhang, Fu, Zhu, Yang, et~al.]{ren2025deepseek}
ZZ~Ren, Zhihong Shao, Junxiao Song, Huajian Xin, Haocheng Wang, Wanjia Zhao, Liyue Zhang, Zhe Fu, Qihao Zhu, Dejian Yang, et~al.
\newblock Deepseek-prover-v2: Advancing formal mathematical reasoning via reinforcement learning for subgoal decomposition.
\newblock \emph{arXiv preprint arXiv:2504.21801}, 2025.

\bibitem[Schulman et~al.(2017)Schulman, Wolski, Dhariwal, Radford, and Klimov]{schulman2017proximalpolicyoptimizationalgorithms}
John Schulman, Filip Wolski, Prafulla Dhariwal, Alec Radford, and Oleg Klimov.
\newblock Proximal policy optimization algorithms, 2017.
\newblock URL \url{https://arxiv.org/abs/1707.06347}.

\bibitem[Shang et~al.(2025)Shang, Wan, Peng, Wu, hui Chen, Yan, and Zhang]{shang2025stepfunprover}
Shijie Shang, Ruosi Wan, Yue Peng, Yutong Wu, Xiong hui Chen, Jie Yan, and Xiangyu Zhang.
\newblock Stepfun-prover preview: Let's think and verify step by step, 2025.
\newblock URL \url{https://arxiv.org/abs/2507.20199}.

\bibitem[Shen et~al.(2025)Shen, Zhang, Huang, Shi, Zhang, Yan, Wang, Wang, Liu, and Lian]{Shen2025DAST}
Yi~Shen, Jian Zhang, Jieyun Huang, Shuming Shi, Wenjing Zhang, Jiangze Yan, Ning Wang, Kai Wang, Zhaoxiang Liu, and Shiguo Lian.
\newblock Dast: Difficulty-adaptive slow-thinking for large reasoning models.
\newblock \emph{arXiv preprint arXiv:2503.04472}, 2025.

\bibitem[Tsoukalas et~al.(2024)Tsoukalas, Lee, Jennings, Xin, Ding, Jennings, Thakur, and Chaudhuri]{tsoukalas2024putnambench}
George Tsoukalas, Jasper Lee, John Jennings, Jimmy Xin, Michelle Ding, Michael Jennings, Amitayush Thakur, and Swarat Chaudhuri.
\newblock Putnambench: Evaluating neural theorem-provers on the putnam mathematical competition.
\newblock \emph{Advances in Neural Information Processing Systems}, 37:\penalty0 11545--11569, 2024.

\bibitem[Wang et~al.(2025)Wang, Unsal, Lin, Baksys, Liu, Santos, Sung, Vinyes, Ying, Zhu, et~al.]{wang2025kimina}
Haiming Wang, Mert Unsal, Xiaohan Lin, Mantas Baksys, Junqi Liu, Marco~Dos Santos, Flood Sung, Marina Vinyes, Zhenzhe Ying, Zekai Zhu, et~al.
\newblock Kimina-prover preview: Towards large formal reasoning models with reinforcement learning.
\newblock \emph{arXiv preprint arXiv:2504.11354}, 2025.

\bibitem[Wenzel et~al.(2008)Wenzel, Paulson, and Nipkow]{wenzel2008isabelle}
Makarius Wenzel, Lawrence~C Paulson, and Tobias Nipkow.
\newblock The isabelle framework.
\newblock In \emph{International Conference on Theorem Proving in Higher Order Logics}, pp.\  33--38. Springer, 2008.

\bibitem[Wu et~al.(2024)Wu, Huang, Zhou, Ying, Wang, Lin, and Chen]{wu2024internlm}
Zijian Wu, Suozhi Huang, Zhejian Zhou, Huaiyuan Ying, Jiayu Wang, Dahua Lin, and Kai Chen.
\newblock Internlm2.5-stepprover: Advancing automated theorem proving via expert iteration on large-scale {LEAN} problems.
\newblock \emph{CoRR}, abs/2410.15700, 2024.
\newblock URL \url{https://arxiv.org/abs/2410.15700}.

\bibitem[Xin et~al.(2024{\natexlab{a}})Xin, Guo, Shao, Ren, Zhu, Liu, Ruan, Li, and Liang]{xin2024deepseekproveradvancingtheoremproving}
Huajian Xin, Daya Guo, Zhihong Shao, Zhizhou Ren, Qihao Zhu, Bo~Liu, Chong Ruan, Wenda Li, and Xiaodan Liang.
\newblock Deepseek-prover: Advancing theorem proving in llms through large-scale synthetic data, 2024{\natexlab{a}}.
\newblock URL \url{https://arxiv.org/abs/2405.14333}.

\bibitem[Xin et~al.(2024{\natexlab{b}})Xin, Ren, Song, Shao, Zhao, Wang, Liu, Zhang, Lu, Du, Gao, Zhu, Yang, Gou, Wu, Luo, and Ruan]{xin2024deepseekproverv15harnessingproofassistant}
Huajian Xin, Z.~Z. Ren, Junxiao Song, Zhihong Shao, Wanjia Zhao, Haocheng Wang, Bo~Liu, Liyue Zhang, Xuan Lu, Qiushi Du, Wenjun Gao, Qihao Zhu, Dejian Yang, Zhibin Gou, Z.~F. Wu, Fuli Luo, and Chong Ruan.
\newblock Deepseek-prover-v1.5: Harnessing proof assistant feedback for reinforcement learning and monte-carlo tree search, 2024{\natexlab{b}}.
\newblock URL \url{https://arxiv.org/abs/2408.08152}.

\bibitem[Xin et~al.(2025)Xin, Xi, Yang, Chen, Wu, Xiao, Sun, Zheng, and Shen]{xin2025bfs}
Ran Xin, Chenguang Xi, Jie Yang, Feng Chen, Hang Wu, Xia Xiao, Yifan Sun, Shen Zheng, and Kai Shen.
\newblock Bfs-prover: Scalable best-first tree search for llm-based automatic theorem proving.
\newblock \emph{CoRR}, abs/2502.03438, 2025.
\newblock URL \url{https://arxiv.org/abs/2502.03438}.

\bibitem[Ying et~al.(2025)Ying, Wu, Geng, Yuan, Lin, and Chen]{ying2025leanworkbooklargescalelean}
Huaiyuan Ying, Zijian Wu, Yihan Geng, Zheng Yuan, Dahua Lin, and Kai Chen.
\newblock Lean workbook: A large-scale lean problem set formalized from natural language math problems, 2025.
\newblock URL \url{https://arxiv.org/abs/2406.03847}.

\bibitem[Zhao et~al.(2024)Zhao, Zheng, Bo, Hu, Thakker, and Kong]{Zhao2024SubgoalXLSE}
Xueliang Zhao, Lin Zheng, Haige Bo, Changran Hu, Urmish Thakker, and Lingpeng Kong.
\newblock Subgoalxl: Subgoal-based expert learning for theorem proving.
\newblock \emph{ArXiv}, abs/2408.11172, 2024.

\bibitem[Zheng et~al.(2022)Zheng, Han, and Polu]{zheng2022miniff}
Kunhao Zheng, Jesse~Michael Han, and Stanislas Polu.
\newblock minif2f: a cross-system benchmark for formal olympiad-level mathematics.
\newblock In \emph{International Conference on Learning Representations}, 2022.
\newblock URL \url{https://openreview.net/forum?id=9ZPegFuFTFv}.

\end{thebibliography}
